\documentclass[10pt, a4paper]{article}

\usepackage{lrec}
\usepackage{graphicx}
\usepackage[utf8]{inputenc}
\usepackage{tabularx}
\usepackage{soul}
\usepackage{epstopdf}
\usepackage[breaklinks=true]{hyperref}
\usepackage{xstring}

\usepackage[usenames,dvipsnames,table,xcdraw]{xcolor}
\usepackage{booktabs}
\usepackage{adjustbox}
\usepackage{enumitem}
\usepackage{contour}
\usepackage[normalem]{ulem}
\usepackage{subcaption}

\contourlength{0.8pt}
\renewcommand{\ul}[1]{%
  \uline{\phantom{#1}}%
  \llap{\contour{white}{#1}}%
}

\definecolor{lila}{rgb}{0.75, 0.25, 0.75}

\newcounter{examples}
\newlist{Example}{enumerate}{10}
\setlist[Example]{label=(\theexamples), align=left, leftmargin=2em}
\let\olditem\item
\newcommand{\exitem}{\olditem \stepcounter{examples}}	
\renewcommand{\item}{\olditem}

\newcommand{\trigger}[1]{\textbf{#1}}
\newcommand{\scope}[1]{\textit{#1}}
\newcommand{\focus}[1]{\ul{#1}}
\newcommand{\dfocus}[1]{\dashuline{#1}}
\newcommand{\eng}[1]{\\ {\footnotesize #1}}

\newcommand{\nubes}{\textsc{\footnotesize NUBes}}

\title{\textsc{NUBes}: A Corpus of Negation and Uncertainty in Spanish Clinical Texts}  

\name{Salvador Lima$^1$, Naiara Perez$^1$, Montse Cuadros$^1$, German Rigau$^2$}

\address{
    $^1$SNLT group at Vicomtech Foundation, Basque Research and Technology Alliance (BRTA),\\ Donostia/San-Sebastián, 20009, Spain\\
    $^2$IXA group. HiTZ centre. University of the Basque Country UPV/EHU, \\Donostia/San-Sebastián, 20018, Spain\\
    \{slima, nperez, mcuadros\}@vicomtech.org, german.rigau@ehu.es\\
 }

\abstract{This paper introduces the first version of the \nubes\ corpus (Negation and Uncertainty annotations in Biomedical texts in Spanish). The corpus is part of an on-going research and currently consists of 29,682 sentences obtained from anonymised health records annotated with negation and uncertainty.  The article includes an exhaustive comparison with similar corpora in Spanish, and presents the main annotation and design decisions. Additionally, we perform preliminary experiments using deep learning algorithms to validate the annotated dataset. As far as we know, \nubes\ is the largest publicly available corpus for negation in Spanish and the first that also incorporates the annotation of speculation cues, scopes, and events. \\ \newline  }

\begin{document}

\maketitleabstract

\section{Introduction}
\label{sec:introduction}

The aim of Natural Language Understanding is to capture the intended meaning of texts or utterances. However, until recently, research has predominantly focused only on propositional aspects of meaning. Truly understanding language involves taking into account many linguistic aspects which are usually overlooked. These linguistic phenomena are sometimes referred to as Extra-Propositional Aspects of Meaning (EPAM) \cite{Morante2012}. Some examples of EPAM include factuality, uncertainty, opinions, beliefs, intentions or subjectivity. Documents enriched with this kind of information can be of utmost importance. For instance, in a domain such as the biomedical, the implicit meaning of a sentence can be crucial to differentiate whether a patient suffers from a disease or not, or whether they should be taking or not a given drug.

One way to learn these nuances is through means of an annotated corpus. Unfortunately, there are not many corpora that cover these phenomena. Just a few consider negation, a key aspect of factuality --and, even fewer, uncertainty. 

Negation is understood as an element that modifies the truth value of an event or a statement, or that makes explicit that an event is absent; uncertainty (also called speculation) occurs when a speaker is not sure whether an event or statement is true. 

Usually, negation and uncertainty are annotated in two parts: on the one hand, the phrase that triggers the change of meaning (called `cue', `trigger' or `marker') and, on the other hand, the words that are affected by them (called `scope'). For a higher level of granularity, there are other elements that can be annotated, such as the element most clearly affected by the cue (called `event'\footnote{In the biomedical domain, `event' refers to any `medical entity', not only to an action, happening, etc.}) or the element that reinforces or diminishes the meaning of the cue (called `polarity'). A typical annotation that includes all these elements is shown in example \ref{ex:intro2}\footnote{In the following examples, cues are marked in bold, scopes in italics, events are underlined and polarity items are enclosed between curly brackets; translations to English are given below each corresponding example.}:

\begin{Example}
    \exitem \label{ex:intro2} La paciente ingresa en UCI con la \trigger{sospecha de} \scope{\{posible\} \focus{encefalitis}} \eng{The patient is admitted to ICU under suspicion of possible encephalitis}
\end{Example}

This paper describes the \nubes\ corpus (Negation and Uncertainty annotations in Biomedical texts in Spanish), a new collection of health record excerpts enriched with negation and uncertainty annotations. To date, \nubes\ is one of the largest available corpus of clinical reports in Spanish annotated with negation, and the first one that includes the annotation of speculation cues, scopes, and events. Additionally, we also present an extension of the IULA-SCRC corpus \cite{Marimon2017} enriched with uncertainty using the same guidelines developed for \nubes. In order to validate the annotated corpus, we present some experimentation using deep neural algorithms. \nubes, the extension of IULA-SCRC --under the name of IULA{\footnotesize+}--, as well as the guidelines, are publicly available\footnote{\href{https://github.com/Vicomtech/NUBes-negation-uncertainty-biomedical-corpus}{https://github.com/Vicomtech/NUBes-negation-uncertainty-} biomedical-corpus}.

The paper is structured as follows: first, a brief survey of related work and corpora is presented. Then, Section \ref{sec:dataset} explains the design decisions, annotation guidelines and the annotation process of \nubes. Next, we present an overview of the main characteristics of the corpus. Section \ref{sec:experiments} presents some preliminary experiments with \nubes. In Section \ref{sec:discussion}, we discuss some of the difficulties faced during the development of the \nubes\ guidelines. Finally, Section \ref{sec:conclusions} presents the conclusions reached and the future work to extend and improve \nubes.

\begin{table*}[!hbtp]
    \centering
    \begin{tabular}{lrrrrrrr}
        \toprule
        & \multicolumn{1}{c}{\shortstack{IxaMed-\\GSC}} & \multicolumn{1}{c}{\shortstack{UHU-\\HUVR}} & \multicolumn{1}{c}{\shortstack{IULA-\\SCRC}} &
        \multicolumn{1}{c}{IULA{\footnotesize+}} & \multicolumn{1}{c}{CL2017} & \multicolumn{1}{c}{\shortstack{Co2017}} & \multicolumn{1}{c}{\nubes} \\
        \midrule
        negation cue & no & yes & yes & yes & yes & yes & yes \\ 
        uncertainty cue & no & no & no & yes & no & yes & yes \\
        scope & no & yes & yes & yes & no & no & yes \\
        event &  yes & yes & no & yes & no & yes & yes \\
        sentences & \footnotesize 5,410 & \footnotesize 8,412 & \footnotesize 3,194\footnotemark[1] & \footnotesize 3,370\footnotemark[1] & ?\footnotemark[2] & ?\footnotemark[3] & \footnotesize 29,682 \\
        \multicolumn{1}{r}{\footnotesize \ \ \ \ \ \ \ \ with negation (\#)} & \footnotesize ? & \footnotesize 2,298 & \footnotesize 1,093 & \footnotesize 957 & \footnotesize ? & \footnotesize ? & \footnotesize 7,567 \\
        \multicolumn{1}{r}{\footnotesize \ \ \ \ \ \ \ \ with negation (\%)} & \footnotesize ?\footnotemark[4] & \footnotesize 27.32 & \footnotesize 34.22 & \footnotesize 28.40 & \footnotesize ? & \footnotesize ?\footnotemark[5]  & \footnotesize 25.49 \\
        \multicolumn{1}{r}{\footnotesize \ \ \ \ \ \ \ \ with uncertainty (\#)} & \footnotesize ? & \footnotesize 0 & \footnotesize 0 & \footnotesize 182 & \footnotesize 0 & \footnotesize ? & \footnotesize 2,219 \\
        \multicolumn{1}{r}{\footnotesize \ \ \ \ \ \ \ \ with uncertainty (\%)} & \footnotesize ?\footnotemark[6] & \footnotesize 0 & \footnotesize 0 & \footnotesize 5.40 & \footnotesize 0 & \footnotesize ? & \footnotesize 7.48 \\
        \bottomrule
    \end{tabular}
    \caption{Comparison between existing biomedical negation and/or uncertainty corpora in Spanish and \nubes, adapted from \protect\newcite{JimenezZafraReview} and \protect\newcite{Marti2018}. \protect\footnotemark[1]\protect\newcite{Marimon2017} report 3,194, but we counted 3,370 sentences in the publicly available corpus. \protect\footnotemark[2]354,677 emergency admission notes. \protect\footnotemark[3]513 radiology reports. \protect\footnotemark[4]27.58\% of the diseases annotated are negated. \protect\footnotemark[5]56\% of the ``findings'' annotated are negated. \protect\footnotemark[6]1.90\% of the diseases annotated are speculative.}
    \label{tab:corpora-comparison}
\end{table*}

\section{Related Work}
\label{sec:related-work}

Negation is such a complex phenomenon that it has been studied from the perspective of multiple fields, ranging from linguistics to philosophy, and even psychology. From a linguistic standpoint, it is a phenomenon that permeates different aspects such as syntax, morphology and semantics. \newcite{stanford-neg} describe it as `an operator [...] that allows for denial, contradiction, and other key properties of human linguistic systems'. Uncertainty is another widely studied topic, as it can also appear in many different settings. It may come from a lack of knowledge or because of how the world is disposed \cite{Kahneman1982VariantsOU}; on top of that, some utterances may only become uncertain within a given context \cite{vincze-2014-uncertainty}.

Due to their significance, their extraction has become a somewhat popular topic in Natural Language Processing \cite{Chapman2001ASA,Huang2007ResearchPA}. This has naturally led to the creation of resources that annotate this phenomena for supervised learning. One of the best known is BioScope \cite{Vincze2008BioScope}, a corpus of biomedical texts in English annotated with both of the previously described phenomena.

In Spanish, seven corpora descriptions have been published for negation. Out of those seven, five are from the clinical domain and only two of them factor in uncertainty: \textit{i)} the IxaMed-GS corpus \cite{Oronoz2015} is a medical texts corpus annotated at an entity-level, that is, some events are characterised as being negated, speculated, or neither; \textit{ii)} the UHU-HUVR corpus \cite{CruzDiaz2017} and \textit{iii)} the IULA Spanish Clinical Record Corpus (IULA-SCRC) \cite{Marimon2017} include negation cues and their scopes; \textit{iv)} \newcite{CampillosLlanos2017} report to be working on extracting negation cue patterns from a corpus of emergency admission notes; finally, \textit{v)} \newcite{Cotik2017} present a corpus of radiology reports annotated with events and relations, including negation and uncertainty. The IxaMed-GS and the corpus by \newcite{Cotik2017} are the only two that annotate uncertainty. Table \ref{tab:corpora-comparison}\footnote{Some of the the articles do not report all the details introduced in the table. Such cases have been marked with a question mark. CL2017 and Co2017 refer to the works by \newcite{CampillosLlanos2017} and \newcite{Cotik2017}, respectively.} provides a general overview of these 5 corpora (plus the two presented in this paper, namely, \nubes\ and IULA{\footnotesize+}). We refer the reader to \newcite{JimenezZafraReview} for a more detailed comparison.

The other two corpora that deal with domains other than the biomedical are the UAM Spanish Treebank \cite{Moreno2003}, a newspaper articles corpus enhanced by \newcite{Moreno2013} to include negation cues and scopes, and the SFU Review\textsubscript{SP}-NEG corpus \cite{Jimenez-Zafra2018sfu}, which studies negation in the context of product reviews.

Of the aforementioned 7 corpora, only UAM Spanish Treebank, SFU Review\textsubscript{SP}-NEG, and IULA-SCRC are publicly available. Thus, to the best of our knowledge, \nubes\ is the second and biggest available corpus in the biomedical domain annotated with negation and uncertainty markers and scopes.

\section{\nubes}
\label{sec:dataset}

\nubes\ derives from a dump of anonymised health records provided by a Spanish private hospital. We extracted plain text from 7 sections consisting of free text --namely, Chief Complaint, Present Illness, Physical Examination, Diagnostic Tests, Surgical History, Progress Notes, and Therapeutic Recommendations--, and split them into sentences with spaCy\footnote{\href{https://spacy.io/}{https://spacy.io/}}. Then, documents were sampled into batches of around 3,000 sentences, by iteratively picking documents from random specialities and sections. 

The anonymisation was done in two steps: first, we annotated manually any item that could be seen as Personal Health Information (PHI), such as names, dates, locations, contact details, and so on. Secondly, we replaced semi-automatically the identified PHI with similar phrases with the help of methods based on rules and dictionaries designed for this purpose \cite{lima+2019b}. As a result, the documents maintain their readability while being suitable for sharing.

All in all, 10 batches have been anonymised and annotated with negation and uncertainty, amounting to 7,019 documents and 29,682 sentences (see Table \ref{tab:corpora-comparison}).


\subsection{Annotation process}
\label{ssec:procedure}

An initial draft of our guidelines was produced by extending IULA-SCRC's to include uncertainty. After annotating IULA-SCRC with this initial draft, we decided to make further changes with respect to negation by annotating \textit{a)} negations inside indirect speech (e.g., `The patient denies'); \textit{b)} verbs that convey a change of state (e.g., `remove'); and, \textit{c)} morphological negation (e.g., `incoherent'). Other minor changes to the guidelines had to be made in order to accommodate uncertainty annotations. These differences with IULA-SCRC and the other corpora are further described in Section \ref{sec:discussion}

After producing the second draft, two linguists worked independently on a first batch of documents of the \nubes\ corpus. Their results were compared and multiple questions and disagreements that arose were discussed. The team also consulted a medical expert who aided them with some difficult scenarios, which are also examined in Section \ref{sec:discussion} All this greatly contributed towards producing the final version of the guidelines.  

Then, the two linguists annotated the same batch adhering to the final guidelines. The inter-annotator agreement was then calculated on the second draft annotations and the final guideline annotations. As Table \ref{tab:iaa} shows, the agreement (Cohen's kappa, $\kappa$, and linearly weighted $\kappa$, $lw\kappa$) improved after the discussion, particularly for cues. The low agreement in polarity items is explained by the fact that they occur very few times (15) and the number of possible tags is also small (2: either it is a polarity item or it is not), which distorts the $\kappa$ measurement. The percentage agreement in the 2$^{nd}$ round for this class is actually 99.95\%.

\begin{table}[!htbp]
    \centering
    \begin{tabular}{lrrrrr}
        \toprule
        &  & \multicolumn{2}{c}{1$^{st}$ round} & \multicolumn{2}{c}{2$^{nd}$ round} \\ 
        & \multicolumn{1}{c}{$N$} & \multicolumn{1}{c}{$\kappa$} & \multicolumn{1}{c}{$lw\kappa$} & \multicolumn{1}{c}{$\kappa$} & \multicolumn{1}{c}{$lw\kappa$} \\
        \midrule
        negation cue & 4 & 0.92 & 0.83 &             \bf{0.93} & \bf{0.89} \\
        uncertainty cue & 3 & 0.81 & 0.81 & \bf{0.84} & \bf{0.84} \\
        scope & 6 & \textbf{0.80} & \textbf{0.76} & \bf{0.80} & \bf{0.76} \\
        event & 6 & 0.79 & 0.74 & \bf{0.80} & \bf{0.75} \\
        polarity item & 2 & 0.40 & 0.40 & \bf{0.50} & \bf{0.50} \\
        all & 14 & 0.82 & 0.77 & \bf{0.83} & \bf{0.79} \\
        \bottomrule
    \end{tabular}
    \caption{Inter-annotator agreement between 2 annotators on the first batch (2,971 sentences). $N$ is the number of tag types considered. The best results are highlighted in bold.}
    \label{tab:iaa}
\end{table}

Finally, a third annotator resolved the differences between the previous two in the first batch in order to create a Gold Standard. Nine more corpus batches and IULA{\footnotesize+} were annotated by one linguist. The current \nubes\ release includes, then, one batch annotated by three people and nine batches produced by a single annotator. We intend to continue working on the corpus and release future versions as we apply the same methodology to the rest of it.

All the annotation work was done with BRAT \cite{stenetorp2012brat}. To speed up the process, an automatic cue annotator service was developed for BRAT  that detects a list of the most frequent cues. On average, we invested around eight hours of annotation work for each batch of $\sim$3,000 sentences.

\subsection{Annotation guidelines}
\label{ssec:guidelines}

\nubes\ includes three main annotated elements: negation cues, uncertainty cues and their scope. Moreover, polarity items and events are also annotated as part of the scope.

\subsubsection{Negation cues}
\label{sssec:negation-triggers}

We define negation cues as elements that modify the truth value of a clause or specify the absence of an entity. Three different types of cues can be distinguished: syntactic, lexical and morphological.

\paragraph{\small Syntactic negation cues.}
\label{par:negsyn}

These are mostly function words or adverbs that can accompany multiple categories of words. It is the simplest type of negation, as well as the most common, as it covers words such as `no' (\textit{no}) and `sin' (\textit{without}):

\begin{Example}
    \exitem \label{ex:NegSynMarker} Fiebre de 38,5 \trigger{sin} \scope{\focus{foco}} \eng{38.5 degrees fever without a focus}
\end{Example}

\noindent Negative time adverbs, such as `nunca' (\textit{never}), can also act as syntactic cues.

\paragraph{\small Lexical negation cues.} 
\label{par:neglex}

They are content words or multi-word expressions that convey negation depending on the context, including verbs, adjectives or noun phrases. These cues are harder to detect as the way in which they negate a phrase is usually subtler than that of syntactic cues. Some examples are `suspender' (\textit{`suspend'}), `incapacidad para' (\textit{`inability to'}) o `descartar' (\textit{`discard'}):

\begin{Example}
    \exitem \label{ex:NegLexMarker} \trigger{Desestiman} \scope{actualmente la realización de \focus{endoscopia}} \eng{At present they dismiss conducting an endoscopy}
\end{Example}

\noindent Noun phrases with negative determiners are also considered lexical cues:

\begin{Example}
    \exitem \label{ex:NegLexNinguna} \trigger{Ninguna de ellas} \scope{de \focus{evolución aguda-subaguda}} \eng{None of them of acute-subacute course}
\end{Example}

\paragraph{\small Morphological negation cues.}
\label{par:morlex}

Morphological negation refers to negation by means of affixes. Since \nubes\ is a medical texts corpus, we decided to limit the annotation of these cues to words that explicitly state the absence of symptoms (`afebril', \textit{afebrile}) or that could be seen as negating a symptom or state (`deshidratado', \textit{dehidrated}). 
Words that do not fulfil those conditions or that are part of a condition name are not annotated. In general, as long as a word could be reformulated as a negated sentence that would be annotated under those conditions, the word would be classified as a cue. For example, `insuficiencia' (\textit{failure}), as in example \ref{ex:no-morph}, was not annotated because `?no suficiencia' is ungrammatical.

\begin{Example}
    \exitem \label{ex:morph} \trigger{Afebril} al ingreso \eng{Afebrile at admission}
    \exitem \label{ex:no-morph} Presentó descompensacion de su insuficiencia cardiaca \eng{[The patient] showed decompensation of their heart failure}
\end{Example}

Finally, it is worth mentioning that not all appearances of negation cues are annotated as such. There are two main cases. On the one hand, there are formulas that simply change the polarity of a positive event without truly negating it (`casi sin', \textit{almost no}; `no siempre', \textit{not always}). These formulas can be restated without the negative cue with no real change in meaning, so we did not consider them. On the other hand, there are negation cues that are actually part of an uncertainty cue, such as `no claro' (\textit{not clear}). This exception will be further developed in the next section.

\subsubsection{Uncertainty cues}
\label{sssec:uncertainty-triggers}

Similarly to negation, uncertainty cues can be separated into two groups: syntactic cues and lexical cues.

\paragraph{\small Syntactic uncertainty cues.} 
\label{par:uncsyn}

Again, these are function words. The only instances of this class are the disjunctions `o' (\textit{or}) and `vs'. These were only annotated when they appeared by themselves in a context of uncertainty \ref{ex:UncertSynMarker}, as they could also appear listing alternatives or as a way to reformulate a sentence or phrase \ref{ex:disjunctions}. 

\begin{Example}
    \exitem \label{ex:UncertSynMarker} \scope{\focus{Una complicación postCNG}} \trigger{o} \scope{\focus{una patología de ori-} \focus{gen digestivo}} \eng{A post-coronary angiography complication or a pathology of digestive origin}
    \exitem \label{ex:disjunctions} En las intercrisis refiere sensación contínua de mareo o inestabilidad \eng{[The patient] mentions continuous dizziness or instability}
\end{Example}

\paragraph{\small Lexical uncertainty cues.} 
\label{par:unclex}

As with lexical negation, these are content words that express uncertainty depending on the context. Some of the most used cues are `probable', `posible' or 'sospecha de' (see \ref{ex:UncertLexMarker1}, \ref{ex:UncertLexMarker2}). Verbs in the conditional mood are also treated as uncertainty cues, including those that usually act as negation cues, as in example \ref{ex:descartaria}.

\begin{Example}
    \exitem \label{ex:UncertLexMarker1} \trigger{Sospecha de} \scope{\focus{dehiscencia de suturas}} \eng{Suspicion of wound dehiscence}
    \exitem \label{ex:UncertLexMarker2} \trigger{Se pensó en} \scope{\focus{un origen funcional de ambos síntomas}} \eng{A functional origin of both symptoms was considered}
    \exitem \label{ex:descartaria} \trigger{Descartaría} \scope{\{de forma razonable\} \focus{una arteritis de} \focus{la temporal} como causa de la clínica} \eng{It would reasonably rule out temporal arteritis as the origin of the symptoms}
\end{Example}

Seemingly negative cues can also express uncertainty depending on the context they appear in. For example, a negated negative cue might be used to express uncertainty \ref{ex:UncertWNeg}, while words that express confidence are also classified as uncertainty when they are negated \ref{ex:NoClaro}. When the latter happens, it might be the case that the cue is discontinuous \ref{ex:DiscMarker}.

\begin{Example}
    \exitem \label{ex:UncertWNeg} \trigger{No se descarta} \scope{\{definitivamente\} \focus{sangrado activo}} \eng{Active bleeding is not definitively ruled out}
    \exitem \label{ex:NoClaro} \trigger{No claro} \scope{\focus{transtorno sensitivo}} \eng{No clear sensitive disorder}
    \exitem \label{ex:DiscMarker} \trigger{Sin} \scope{signos} \trigger{claros} \scope{de \focus{isquemia aguda}} \eng{No clear signs of severe ischemia}
\end{Example}

Finally, negation can also happen together with uncertainty in the same sentence. There are two possible scenarios. If an uncertainty cue appears within the scope of a negation, the latter usually invalidates the meaning of the uncertainty. For example, in \ref{ex:invalidatedUnc}, `sugestiva de' stops indicating that the speaker is unsure of what they say when it is negated by `no'. In such cases, the uncertainty cue is not annotated. However, if an uncertainty cue is the one that appears first and scopes over a negation cue, the meaning is maintained, as in example \ref{ex:negInsideUnc}\footnote{In this example, the scope of the embedded cue is marked with a dotted underline.}. This time, the negation cue as well as its own scope are annotated inside the uncertainty's scope.

\begin{Example}
    \exitem \label{ex:invalidatedUnc} \trigger{No} \scope{refiere clínica sugestiva de \focus{aura migrañosa}} \eng{[The patient] does not allude to symptoms suggestive of migraine aura}
    \exitem \label{ex:negInsideUnc} \trigger{Sospecha de} \scope{\{posible\} \focus{HSA} \trigger{no} \dfocus{apreciada en el TAC}} \eng{Suspicion of a possible subarachnoid hemorrhage not detected in the CT}
\end{Example}

\subsubsection{Scopes}
\label{sssec:scopes}

The \textbf{scope} is the part of the sentence whose meaning is changed by a negation or uncertainty cue. We follow IULA-SCRC's definition of the scope as ``the maximal syntactic unit that is affected by the marker'' \cite[p. 46]{Marimon2017}. 
In \nubes, coordinated items are included within the scope, but cues are not. Subjects are only included when they appear in post-verbal position. 

As the scope is always the maximal syntactic unit, it is sometimes longer than the actual part that is most prominently affected by negation or uncertainty. For that reason, we also annotate \textbf{events} inside the scope, as in \ref{ex:focus}: 

\begin{Example}
    \exitem \label{ex:focus} \trigger{No} \scope{se aprecian \focus{lesiones estructurales}} \eng{No structural lesions are observed}
\end{Example}

When a sentence contains multiple noun phrases coordinated, each of them is annotated as an individual event inside a bigger scope, as in example \ref{ex:focusCoord}. However, if the modifiers of the same noun phrase are separated by coordination, they are still all treated as part of the same event \ref{ex:focusCoord2}:

\begin{Example}
    \exitem \label{ex:focusCoord} \trigger{Sin aparente} \scope{\focus{TCE} ni \focus{focalidad}} \eng{With no apparent TBI or [neurological] focus}
    \exitem \label{ex:focusCoord2} \trigger{No} \scope{\focus{clínica digestiva ni miccional}} \eng{No digestive nor voiding symptoms}
\end{Example}

Events are labelled with a set of medical entity tags adapted from IULA-SCRC's interpretation of the SNOMED-CT classification\footnote{\href{http://www.snomed.org/}{http://www.snomed.org/}}: Medical findings and Disorders, Medical Procedures, Chemicals and Body Substances, Body Structure, Other --for medical concepts outside of the previous categories; not in IULA-SCRC-- and Phrase --used for general scopes and entities outside of the medical field. If the event and the scope match in span, the most specific label is used for the whole scope. Otherwise, the event is annotated inside a longer Phrase label.

The scope of a cue can sometimes be \textbf{discontinuous}. That is, a cue can affect multiple text spans that are separated. The most frequent structures that trigger discontinuous scopes are the following: \textit{a)} the cue appears within the affected phrase \ref{ex:DiscScope1}, causing the cue to be surrounded by its scope; \textit{b)} the object of a verb has been omitted or substituted by a pronoun --in \ref{ex:DiscScope2}, ``them'' substitutes ``inhalers'' and thus the latter is annotated as being part of the scope; \textit{c)} there is ellipsis of the verb, as in \ref{ex:DiscScope3}, where the verb ``repeats'' is omitted in the second sentence as it has already been used before. Thus, the first mention is annotated as being part of the scope. Discontinuous scopes also happen frequently in combination with discontinuous cues, as in example \ref{ex:DiscScMark}.

\begin{Example}
    \exitem \label{ex:DiscScope1} \scope{Relación} \trigger{probable} \scope{con \focus{incipientes cambios por otitis} \focus{media crónica}} \eng{Probable relation to early changes caused by chronic otitis media}
    \exitem \label{ex:DiscScope2} Refiere su Médico de Cabecera que le pautó \scope{\focus{inhaladores}} pero \trigger{no} los \scope{tolera} \eng{Her family doctor refers that she gave him inhalers but he does not tolerate them}
    \exitem \label{ex:DiscScope3} \scope{Repite} palabras sencillas pero \trigger{no} \scope{\focus{frases}} \eng{[The patient] repeats simple words but not sentences}
    \exitem \label{ex:DiscScMark} \trigger{No pudiendo precisar si} \scope{ha presentado} \trigger{o no} \scope{\focus{pérdida de conciencia}} \eng{[The patient] is not able to specify whether they lost consciousness or not}
\end{Example}

Finally, elements expressing \textbf{polarity} changes can also appear inside the scope. They are elements that reinforce the expressive power of the phenomena. Usually, these are pronouns or negative determiners, such as `alguna' or `ninguna' (\textit{any}), but multiple cues of the same type appearing together are also treated as such if they were used to reaffirm the meaning of the first cue.
\begin{Example}
    \exitem \label{ex:polaritypronouns} \trigger{Niega} \scope{\focus{dolor a \{ningún\} nivel}} \eng{[The patient] denies pain at any level}
    \exitem \label{ex:polaritymarkers} \trigger{Parece} \scope{detectarse un \{posible\} \focus{deterioro cognitivo} \focus{de \{posible\} origen vascular}} \eng{A possible cognitive impairment of possible vascular origin has seemingly been detected}
\end{Example}

Nevertheless, there are some rare cases where it is also possible for negation or uncertainty to appear embedded inside another negation cue's scope without actually reinforcing the meaning of the first cue (e.g. if the second cue is part of a modifier of the negated event or the scope of the first cue is a long embedded clause). In such cases, both cues and their scope are annotated separately, as in example \ref{ex:nopol}.

\begin{Example}
    \exitem \label{ex:nopol} \trigger{Imposibilidad para} \scope{una bipedestación \trigger{sin} \dfocus{ayuda}} \eng{Inability to stand without help}
\end{Example}

\subsection{Dataset statistics}
\label{ssec:statistics}

\begin{table}[!htbp]
    \centering
    \begin{tabular}{lr}
        \toprule
        documents & 7,019 \\
        sentences & 29,682 \\
        tokens & 518,068 \\
        vocabulary size & 31,698 \\
        \midrule
        \multicolumn{2}{l}{\textit{negation}} \\
        \midrule
        sentences affected & 7,567 \\
        average cues per affected sentence & 1.25 $\pm$ 0.66 \\
        discontinuous cues & 0 \\
        average scope size in tokens & 4.01 $\pm$ 3.59 \\
        discontinuous scopes & 219 \\
        \midrule
        \multicolumn{2}{l}{\textit{uncertainty}} \\
        \midrule
        sentences affected & 2,219 \\
        average cues per affected sentence & 1.12 $\pm$ 0.38 \\
        discontinuous cues & 95 \\
        average scope size in tokens & 5.27 $\pm$ 4.97 \\
        discontinuous scopes & 123 \\
        \bottomrule
    \end{tabular}
    \caption{Size of \nubes}
    \label{tab:dataset-size}
\end{table}

The part of the  corpus that has been annotated so far consists of 29,682 sentences, out of which 7,567 (25.49\%) include negation and 2,219 (7.48\%) include uncertainty. A general overview of the size of \nubes\ is described in Table \ref{tab:dataset-size}. In many of the sentences there is more than one cue, and both phenomena might appear together and/or independently. Discontinuous cues and scopes seem to be much more frequent for uncertainty than for negation.

The distribution of both phenomena over the different medical report sections follows the same pattern. Unsurprisingly, negation and uncertainty are more frequent in sections that tend to be longer (i.e., Progress Notes, Diagnostic Tests, and Present Illness). Their distribution does not fit into the same pattern, however, when analysed over medical specialities. Neurology reports stand out in particular for their high usage of speculative expressions. Negation, on the other hand, is most frequent in Cardiology, General Surgery, Neurology, and Internal Medicine.

Concerning the different cues that appear in the corpus, 345 unique negation and 297 unique uncertainty cues have been annotated. The most frequent cues sorted by type are shown in Table \ref{tab:top-triggers}.

\begin{table}[!htpb]
    \centering
    \begin{tabular}{lr}
        \toprule
        & freq. \\
        \midrule
        \multicolumn{2}{l}{\textit{syntactic negation}} \\
        \midrule
        no (\textit{no}, \textit{not}) & 4,058 \\ 
        sin (\textit{without}) & 2,518 \\ 
        tampoco (\textit{neither}) & 40 \\
        nunca (\textit{never}) & 5 \\
        excepto (\textit{except}) & 4 \\
        \midrule
        \multicolumn{2}{l}{\textit{lexical negation}} \\
        \midrule
        negativo (\textit{negative}, sg.) & 123 \\ 
        negativos (\textit{negative}, pl.) & 99 \\ 
        retirada de (\textit{withdrawal of}) & 96 \\ 
        niega (\textit{denies}) & 83 \\ 
        suspender (\textit{withhold}) & 59 \\ 
        \midrule
        \multicolumn{2}{l}{\textit{morphological negation}} \\
        \midrule
        afebril (\textit{afrebile}) & 252 \\ 
        asintomático (\textit{asymptomatic}, m.) & 241 \\ 
        asintomática (\textit{asymptomatic}, f.) & 150 \\ 
        inespecífico (\textit{non-specific}) & 39 \\ 
        asintomatico (sic) & 34 \\ 
        \midrule
        \multicolumn{2}{l}{\textit{syntactic uncertainty}} \\
        \midrule
        vs & 13 \\
        o (\textit{or}) & 4 \\
        versus & 1 \\
        vs. & 1 \\
        \midrule
        \multicolumn{2}{l}{\textit{lexical uncertainty}} \\
        \midrule
        probable & 357 \\
        posible (\textit{possible}) & 198 \\
        compatible con (\textit{compatible with}) & 188 \\
        sospecha de (\textit{suspicion of}) & 144 \\
        parece (\textit{seems}) & 130 \\ 
        \bottomrule
    \end{tabular}
    \caption{The 5 most common cues by type}
    \label{tab:top-triggers}
\end{table}

\section{Experiments with \nubes}
\label{sec:experiments}

A set of experiments have been conducted in order to ascertain the validity of \nubes\ and establish a competitive baseline on this corpus. The task evaluated has been automatic negation and uncertainty cue and scope labelling with the BIO scheme \cite{Ramshaw1999}\footnote{Please note that the aim of this work is not to find the best possible algorithm or features for automatic negation detection in Spanish. For literature specific to the topic, please refer to \newcite{Santiso2019}, \newcite{Loharja2018}, \newcite{Fabregat2018} or \newcite{koza2019automatic}, among others.}.

\subsection{Data}
\label{ssec:exp-dataset}

The dataset used contains all the sentences with at least one negation or uncertainty annotation (9,202) plus as many sentences with no annotations whatsoever. This dataset has been shuffled and split into train (75\%), development (10\%), and test (15\%) sets. Table \ref{tab:experiment-data} shows the size of these splits. Marker and scope labels have been simplified to 4 generic categories: negation or uncertainty marker, and negation or uncertainty scope. Scopes have been flattened to the biggest span possible, thus ignoring events, coordination and polarity particles.

\subsection{Methodology}
\label{ssec:exp-method}

We use NCRF{\footnotesize ++} \cite{yang2018ncrf}, an open-source toolkit built upon PyTorch to develop neural sequence labelling architectures. Out-of-the-box network configuration\footnote{4 CNN layers of 50 dimensions for character sequence representations, a biLSTM layer of 200 dimensions for word sequence representations, and an output CRF layer; see \href{https://github.com/jiesutd/NCRFpp/blob/master/readme/Configuration.md}{https://github.com/jiesutd/NCRFpp}.} and hyperparameters have been kept, except for the batch size (16), the learning rate (00.5) and learning rate decay (00.1). Several groups of input features at token level have been tested, namely:
\begin{itemize}[noitemsep]
    \item \textbf{form}: affixes of 2 and 3 characters, and whether the token is a punctuation mark, a number or an alphabetic string.
    \item \textbf{morphsyn}: the token's lemma, its part-of-speech tag, the type of dependency relation, and the lemmas of the dependent children to the right and left, all extracted with spaCy's es-core-news-md 2.2.0 model.
    \item \textbf{brown}: Brown cluster \cite{brown1992class} complete paths and paths pruned at lengths 16, 32, and 64. The clusters were learned with tan-clustering\footnote{\href{https://github.com/mheilman/tan-clustering}{https://github.com/mheilman/tan-clustering}} from the training set and the 11,278 sentences left out from the dataset split.
    \item \textbf{metadata}: the speciality and section the sentence has been extracted from.
    \item \textbf{window}: all the features of the neighbouring tokens in a $\pm$ 2 window.
\end{itemize}

\begin{table}[!tbp]
    \centering
    \begin{tabular}{lrrr}
        \toprule
        & train & dev. & test \\
        \midrule
        sentences & 13,802 & 1,840 & 2,762 \\
        negation cues & 6,976 & 919 & 1,423 \\
        negation scopes & 6,379 & 847 & 1,322 \\
        uncertainty cues & 1,866 & 263 & 400 \\
        uncertainty scopes & 1,886 & 260 & 400 \\
        \bottomrule
    \end{tabular}
    \caption{Size of the corpus subset used in the experiments}
    \label{tab:experiment-data}
\end{table}

An ablation study has been performed by withdrawing one group of features each time. We have also trained a model with just the tokens as features. In total, then, 7 systems have been trained: one with all the features available, another with just tokens as features, and one per --ablated-- feature group. For each system, we have kept the model that has obtained the best F1-score against the development split within 40 epochs.

The experiment has been run 5 times with random seed initialisation. We report the mean and standard deviation of the micro-averaged precision, recall and F1-score of the 5 runs for each system. We have also computed the statistical significance per the Bootstrap test \cite{efron1994introduction} of \textit{a)} the differences between the results of each run, and \textit{b)} the difference between each ablation model and the base model that uses all the features available.

\subsection{Results}
\label{ssec:exp-results}

The results of the negation and uncertainty detection are shown in Tables \ref{tab:ncrfpp-negation} and \ref{tab:ncrfpp-uncertainty}, respectively.

\begin{table*}[!htbp]
    \centering
    
    \begin{subtable}{\textwidth}
    \centering
    \begin{tabular}{lrrrrrrrrrrrr}
        \toprule
        & \multicolumn{3}{c}{negation marker} & \multicolumn{3}{c}{negation scope} \\
        \midrule
        & \multicolumn{1}{c}{P} & \multicolumn{1}{c}{R} & \multicolumn{1}{c}{F1} & \multicolumn{1}{c}{P} & \multicolumn{1}{c}{R} & \multicolumn{1}{c}{F1} \\
        \midrule
        \footnotesize all features                  & 96.1 $\pm$ 0.3 & 95.0 $\pm$ 0.3 & 95.5 $\pm$ 0.1 & 93.0 $\pm$ 0.9 & 88.3 $\pm$ 1.0 & 90.6 $\pm$ 0.4 \\
        \midrule
        \multicolumn{1}{r}{\footnotesize -form}     & $^{\dagger}$96.2 $\pm$ 0.3 & $^{\dagger}$94.8 $\pm$ 0.2 & $^{\dagger}$95.5 $\pm$ 0.1 & $^{\dagger}$93.0 $\pm$ 1.7 & $^{*,\dagger}$88.0 $\pm$ 1.2 & $^{\dagger}$90.4 $\pm$ 0.3 \\
        \multicolumn{1}{r}{\footnotesize -morphsyn} & $^{\dagger}$95.9 $\pm$ 0.4 & $^{\dagger}$95.4 $\pm$ 0.1 & $^{\dagger}$95.6 $\pm$ 0.1 & $^{\dagger}$92.5 $\pm$ 1.9 & $^{*,\dagger}$87.9 $\pm$ 1.5 & $^{\dagger}$90.1 $\pm$ 0.4 \\
        \multicolumn{1}{r}{\footnotesize -brown}    & $^{\dagger}$96.1 $\pm$ 0.2 & $^{\dagger}$95.3 $\pm$ 0.2 & $^{\dagger}$95.7 $\pm$ 0.2 & $^{\dagger}$92.9 $\pm$ 0.9 & $^{*,\dagger}$88.1 $\pm$ 0.7 & $^{\dagger}$90.5 $\pm$ 0.2 \\
        \multicolumn{1}{r}{\footnotesize -metadata} & $^{\dagger}$96.1 $\pm$ 0.2 & $^{\dagger}$95.3 $\pm$ 0.2 & $^{\dagger}$95.7 $\pm$ 0.1 & $^{\dagger}$93.6 $\pm$ 0.6 & $^{*,\dagger}$87.4 $\pm$ 0.4 & $^{\dagger}$90.4 $\pm$ 0.1 \\
        \multicolumn{1}{r}{\footnotesize -window}   & $^{*,\dagger}$96.1 $\pm$ 0.4 & $^{*}$94.7 $\pm$ 0.2 & $^{*}$95.4 $\pm$ 0.2 & 93.1 $\pm$ 0.8 & $^{*}$88.0 $\pm$ 0.9 & $^{*}$90.5 $\pm$ 0.4 \\
        \midrule
        \multicolumn{1}{l}{\footnotesize tokens}    & $^{\dagger}$96.5 $\pm$ 0.2 & 94.5 $\pm$ 0.3 & $^{\dagger}$95.5 $\pm$ 0.1 & 92.0 $\pm$ 0.6 & $^{*}$86.2 $\pm$ 0.4 & 89.0 $\pm$ 0.2 \\
        \bottomrule
    \end{tabular}
    \caption{Results of negation marker and scope recognition and classification}
    \label{tab:ncrfpp-negation}
    \end{subtable}
    
    \vspace{\baselineskip}
    
    \begin{subtable}{\textwidth}
    \centering
    \begin{tabular}{lrrrrrrrrrrrr}
        \toprule
        & \multicolumn{3}{c}{uncertainty marker} & \multicolumn{3}{c}{uncertainty scope} \\
        \midrule
        & \multicolumn{1}{c}{P} & \multicolumn{1}{c}{R} & \multicolumn{1}{c}{F1} & \multicolumn{1}{c}{P} & \multicolumn{1}{c}{R} & \multicolumn{1}{c}{F1} \\
        \midrule
        \footnotesize all features                  & 86.9 $\pm$ 1.0 & 83.2 $\pm$ 0.6 & 85.0 $\pm$ 0.3 & 83.4 $\pm$ 2.6 & 74.4 $\pm$ 3.4 & 78.5 $\pm$ 0.7 \\
        \midrule
        \multicolumn{1}{r}{\footnotesize -form}     & $^{\dagger}$87.6 $\pm$ 2.2 & $^{\dagger}$82.1 $\pm$ 1.8 & $^{\dagger}$84.7 $\pm$ 0.8 & $^{\dagger}$85.3 $\pm$ 1.8 & 71.3 $\pm$ 3.4 & $^{\dagger}$77.6 $\pm$ 1.4 \\
        \multicolumn{1}{r}{\footnotesize -morphsyn} & $^{\dagger}$86.9 $\pm$ 1.2 & $^{\dagger}$83.4 $\pm$ 1.3 & $^{\dagger}$85.1 $\pm$ 1.1 & $^{\dagger}$83.6 $\pm$ 1.2 & $^{\dagger}$73.3 $\pm$ 1.0 & $^{\dagger}$78.1 $\pm$ 0.5 \\
        \multicolumn{1}{r}{\footnotesize -brown}    & $^{\dagger}$86.9 $\pm$ 1.1 & $^{\dagger}$82.4 $\pm$ 0.6 & $^{\dagger}$84.6 $\pm$ 0.7 & $^{\dagger}$83.2 $\pm$ 1.5 & $^{\dagger}$73.5 $\pm$ 1.3 & $^{\dagger}$78.0 $\pm$ 0.9 \\
        \multicolumn{1}{r}{\footnotesize -metadata} & $^{\dagger}$88.3 $\pm$ 1.2 & $^{\dagger}$82.3 $\pm$ 0.5 & $^{\dagger}$85.2 $\pm$ 0.5 & $^{*.\dagger}$86.7 $\pm$ 2.0 & $^{*}$72.4 $\pm$ 1.9 & $^{*,\dagger}$78.8 $\pm$ 0.4 \\
        \multicolumn{1}{r}{\footnotesize -window}   & $^{*,\dagger}$86.5 $\pm$ 1.1 & $^{*}$81.6 $\pm$ 1.3 & $^{*}$84.0 $\pm$ 0.5 & $^{*,\dagger}$81.9 $\pm$ 2.6 & $^{*}$72.2 $\pm$ 3.7 & $^{*}$76.6 $\pm$ 1.4 \\
        \midrule
        \multicolumn{1}{l}{\footnotesize tokens} & 86.4 $\pm$ 2.0 & $^{*}$79.8 $\pm$ 1.1 & $^{*}$82.9 $\pm$ 0.5 & $^{*}$80.7 $\pm$ 2.8 & $^{*}$69.5 $\pm$ 2.1 & 74.6 $\pm$ 0.3 \\
        \bottomrule
    \end{tabular}
    \caption{Results of uncertainty marker and scope recognition and classification}
    \label{tab:ncrfpp-uncertainty}
    \end{subtable}
    
    \caption{Results of experiments; $^{*}$the differences between the results obtained by this model in the 5 runs \textit{are} statistically significant with p-value $<$ 0.05; $^{\dagger}$the difference w.r.t. using all the features \textit{is not} significant with p-value $>$ 0.05}
    \label{tab:ncrfpp}
    
\end{table*}

Overall, a sharp difference between negation and uncertainty detection can be observed. Unsurprisingly, negation detection seems to be an easier task than uncertainty detection --marker F1-score 95.0 vs 83.2; scope F1-score 90.6 vs 78.5--, which can be explained by the fact that we have more examples of the former case. Moreover, speculation cues and scopes are more likely to be discontinuous and the variability of speculation cues is also higher, which adds difficulty to their correct identification.

Regarding the impact of the different groups of features, removing individual groups seems to have little effect, the differences not being statistically significant in most of the occasions. Notwithstanding, removing \textit{all} features yields significantly worse results for all categories except negation marker detection. The difference is sharper, again, for uncertainty marker and scope detection, which seems to benefit more from the features, particularly of window features.

All in all, the experiments show that the corpus is useful for training models for negation and speculation detection. Nevertheless, there is ample room to improve the results, specially of uncertainty detection.

\subsection{Error analysis}
\label{ssec:exp-error}

Most of the errors involve post-scope and discontinuous markers. Although the cues \textit{are} properly detected, their preceding scopes are not. This happens in a variety of structures, such as relative clauses \ref{ex:err1}, postnominal adjectives \ref{ex:err2}, and phrases formatted with colons \ref{ex:err3}. Note that the examples in this section show \textit{incorrect} annotations made by the trained taggers:

\begin{Example}
    \exitem \label{ex:err1} Sangrado que \trigger{desaparece} \eng{Bleeding that goes away}
    \exitem \label{ex:err2} Control en heces \trigger{negativo} \eng{Negative stool test}
    \exitem \label{ex:err3} Examen anatomopatológico: \trigger{no} \eng{Anatomopathological examination: no}
\end{Example}

\noindent This type of error is reduced somewhat when exploiting all the features described. Moreover, additional experiments without the biLSTM architecture have proven that it is beneficial in this regard, although insufficient. 

Another frequent error arises from the incorrect capitalisation and/or punctuation in the input texts, which leads to scopes ranging beyond sentence boundaries. 

Co-ocurrence of several cues --of the same \ref{ex:err4} or different \ref{ex:err5} type-- within a short text span also introduces errors. Uncertainty markers starting with `no' or `sin' \ref{ex:err6} are particularly tricky.

\begin{Example}
    \exitem \label{ex:err4} \trigger{Parece} \scope{poco} \trigger{probable} \scope{que [...]} \eng{It seems hardly likely that [...]}
    \exitem \label{ex:err5} \trigger{No parece} \scope{identificarse ningún factor} \eng{No factor seems to be identified}
    \exitem \label{ex:err6} \trigger{Sin} \scope{focos claros} \eng{No clear foci}
\end{Example}

Finally, sources of less common errors include: markers that occur seldom in the corpus and thus are hard to detect automatically; long and complex scopes that involve coordination and/or subordination of several clauses; and the inclusion of a preposition or complementiser as being part of the marker instead of the scope, or vice versa.

\section{Discussion}
\label{sec:discussion}

In this section, we report some of the main difficulties faced during the creation of the \nubes\ corpus. Annotating a corpus with extra-propositional meaning requires a thorough linguistic analysis that led to many discussions before, during and even after the process. Aspects like how to demarcate the definition of negation and uncertainty and whether some examples were actually part of them proved to be a source of disagreement. On top of that, the idiosyncrasies of medical language also posed some complications, mostly vocabulary-related.

The first step of the corpus creation process was to reach an agreement on what the terms negation and uncertainty encompass. An overview of the existing literature both in English and Spanish, revealed the there is not a clear-cut definition of the phenomenon across corpora. As a consequence, each corpus has been annotated with a different criterion. The main differences between them have to do with what is accepted as negation and the way in which elements such as scope are annotated.



We ultimately considered that our definition of negation (see Section \ref{sec:introduction}) should also encompass every word that implies that an entity is not occurring or has not occurred: either at all (`imposibilidad para', (\textit{impossibility to})) or anymore (`retirada de' (\textit{removal of}), `suspender' (\textit{withhold})). Some authors such as \newcite{Marimon2017} argue that they did not take into account the cues we have just mentioned because they express a ``change of state'' (ibid.) or, in the case of `negar' (\textit{deny}), that it ``is considered, in factual terms, an statement of what someone says''. 
However, we consider that figuring out whether a statement is actually indirect speech or not is a different task. 

Another debatable example is the postnominal adjective `negativo' (\textit{negative}). The authors of UHU-HUVR \cite{CruzDiaz2017} only annotate this word for test results whenever the name of the test and that of the condition is the same, as it means that the patient does not have said condition; otherwise, it means that the test has taken place and the results are simply negative. This contrast is shown respectively in examples \ref{ex:uhu1} and \ref{ex:uhu2}, taken from UHU-HUVR.

\begin{Example}
    \exitem \label{ex:uhu1} Serología materna: [Toxoplasma]: Negativo \eng{Maternal serology: Toxoplasma: Negative}
    \exitem \label{ex:uhu2} Técnicas de Z-N (normal y largo) negativo \eng{Negative Z-N stain (normal and long)}
\end{Example}

\noindent In \nubes, the latter case \ref{ex:uhu2} is also annotated as it still accommodates into our definition of negation. 
The only exception is when `negativo' is part of an entity's name, e.g. `bacterias Gram negativas' (\textit{Gram-negative bacteria}).

In spite of our broad definition of negation, not all negative occurrences have been annotated. Negative polarity verbs have only been considered when they appear in performative utterances. That is, conditional constructions \ref{ex:conditional}, volition verbs \ref{ex:wishes} or final adjuncts \ref{ex:para} have not been annotated:


\begin{Example}
    \exitem \label{ex:conditional} Si fiebre alta que no cede [...] \eng{If [they have] high fever that doesn't drop [...]}
    \exitem \label{ex:wishes} Refiere molestias y quiere quitárselo \eng{[The patient] says it hurts and wants it removed}
    \exitem \label{ex:para} Varón de 68 años, remitido desde su C.Salud, para descartar TVP \eng{68-year-old male sent by their local clinic to discard DVT}
\end{Example}

\noindent However, this rule requires considering the context of the statement. For example, conditionals can take place next to an uncertainty cue (as in example \ref{ex:DiscScMark}, repeated here for convenience as \ref{ex:conditional_ann}, or example \ref{ex:conditional_ann_2}, where the conditional form only reinforces the uncertainty), or a final adjunct might refer to an event that has already taken place \ref{ex:para_ann}. These special cases \textit{are} annotated. 


\begin{Example}
    \exitem \label{ex:conditional_ann} \trigger{No pudiendo precisar si} \scope{ha presentado} \trigger{o no} \scope{\focus{pérdida de conciencia}} \eng{[The patient] is not able to specify whether they lost consciousness or not}
    \exitem \label{ex:conditional_ann_2} Sugerimos una valoración psiquiatrica, \trigger{por si} \scope{el origen del cuadro pudiera estar generado o influenciado por un cuadro depresivo} \eng{We suggest a psychiatric evaluation, in case the symptoms could be generated or influenced by a depressive disorder}
    \exitem \label{ex:para_ann} Ingresa para \trigger{retirada de} \scope{\focus{infusor} \focus{quimioterapico} el 21/03/09} \eng{[The patient] is admitted on 21/03/09 for chemotherapy infuser removal.}
\end{Example}


Uncertainty also posed some difficulties due to the general vagueness of the medical field and the use of medical jargon. Because of this, we had problems determining whether some expressions were negative, speculative, or neither. A medical practitioner assisted the final decisions in these cases. 
Some of the most compelling cases include: `orientar(se)' (lit: \textit{be oriented as}) \ref{ex:orienta}, which at first we interpreted as conveying an assertion, but it actually has a layer of uncertainty; `asociar' (lit: \textit{associate}) \ref{ex:asocia} seemed like it could express uncertainty depending on the context, but it is just used to state the co-occurrence of several symptoms or diseases; `impresionar' (lit: \textit{to impress, to move}) \ref{ex:impresionade}, from `dar la impresión de' (\textit{strike as}, \textit{look like}), is a commonly used verb to convey uncertainty, although it only has this meaning in the medical domain.

\begin{Example}
    \exitem \label{ex:orienta} Todo ello \trigger{orienta} \scope{junto con la clínica a \focus{un cuadro} \focus{suboclusivo}} \eng{All this, along with the symptoms, points out to a subocclusion case}
    \exitem \label{ex:asocia} Tras limpieza quirúrgica se asocia al tto con antifúngicos \eng{After surgical cleaning, it is associated to the antifungal treatment}
    \exitem \label{ex:impresionade} [...] presentando la exploración descrita \trigger{impresionando} \scope{el cuadro de \focus{síndrome} \focus{confusional}} \eng{[...] resulting the exploration as described, the case \textit{impressing as} a confusional state}
\end{Example}


Interestingly, these expressions are difficult to classify because they express uncertainty at different levels (e.g. from \textit{almost certain} to \textit{completely unsure}). We are considering expanding the annotations to include the different levels of confidence and uncertainty in the future to better deal with these cases.

Finally, some of the instances that are categorised as negation by other corpora were annotated as uncertainty in \nubes\ due to the inclusion of this phenomenon. For example, given the sequence `sin clara' (\textit{no clear}), IULA-SCRC annotates `sin' as a cue and `clara' as part of the scope. In \nubes, `sin clara' as a whole is considered an uncertainty cue.

\section{Conclusions and Future Work}
\label{sec:conclusions}

In this paper we have presented the \nubes\ corpus, a new collection of biomedical texts in Spanish annotated for negation and uncertainty. As far as we know, \nubes\ is the largest corpus of clinical reports in Spanish annotated with negation and the first one including the annotation of speculation cues, scopes, and events. We have explored the corpus from different perspectives: by its comparison with similar corpora, by justifying its design and by explaining the challenges faced during its creation. Furthermore, preliminary experiments have been conducted with the corpus in order to ascertain its validity and establish a competitive baseline. \nubes, IULA{\footnotesize+}, as well as the annotation guidelines are publicly available from the web\footnote{\href{https://github.com/Vicomtech/NUBes-negation-uncertainty-biomedical-corpus}{https://github.com/Vicomtech/NUBes-negation-uncertainty-} biomedical-corpus}.

As part of on-going work, we expect to improve the quality of \nubes. At the moment, $\sim$10\% of the corpus has been annotated by three people, while the rest has been produced by a single annotator. Another line of future work includes performing different and more exhaustive experiments with \nubes, such as testing other sequence labelling algorithms and architectures, or exploiting the relations between cues and scopes.

\section{Acknowledgements}

This work has been supported by Vicomtech and partially funded by the project DeepReading (RTI2018-096846-B-C21, MCIU/AEI/FEDER,UE).

\section{Bibliographical References}
\label{main:ref}

\bibliographystyle{lrec}
\bibliography{lrec2020nubes}

\end{document}